\documentclass[10pt,twocolumn,letterpaper]{article}
\usepackage{graphicx}
\usepackage{amsmath}
\usepackage{amssymb}
\usepackage{booktabs}
\usepackage{enumitem}
\usepackage{verbatim}

\usepackage{subfigure}

\usepackage{wrapfig}
\usepackage{times}
\usepackage{epsfig}
\usepackage[utf8]{inputenc} % allow utf-8 input
\usepackage[T1]{fontenc}    % use 8-bit T1 fonts
\usepackage[pagenumbers]{cvpr} % To force page numbers, e.g. for an arXiv version
\usepackage[pagebackref,breaklinks,colorlinks]{hyperref}
\usepackage[capitalize]{cleveref}
\usepackage{siunitx}
\usepackage{cite}
\usepackage{ctable}
\usepackage{hhline}
\usepackage{booktabs}
\usepackage{caption}
\usepackage{csquotes}
\usepackage{pifont}
\usepackage{mathtools}
\usepackage{animate}

\usepackage[vlined,ruled]{algorithm2e}
\usepackage{ctable}
\usepackage{soul}
\usepackage{multicol}
\usepackage{makecell}
\usepackage[normalem]{ulem}
\usepackage{multirow}
\usepackage{float}
\usepackage{stfloats}
\usepackage{colortbl} % for cell colors
\usepackage{xcolor}  % for color definitions
\usepackage{stmaryrd}
\usepackage{array}
\usepackage{xspace}
\usepackage{multirow}
\usepackage{threeparttable} 
\usepackage{subfigure} 
\newcommand*{\affaddr}[1]{#1} % No op here. Customize it for different styles.
\newcommand*{\affmark}[1][*]{\textsuperscript{#1}}

\title{DeCoP: Enhancing Self-Supervised Time Series Representation with Dependency Controlled Pre-training}
\makeatletter
\newcommand{\printfnsymbol}[1]{%
  \textsuperscript{\@fnsymbol{#1}}%
}
\makeatother

\author{%
Yuemin Wu\affmark[1], Zhongze Wu\affmark[2], Xiu Su\affmark[2], Feng Yang\affmark[3], Hongyan Xu\affmark[2], Xi Lin\affmark[4], Wenti Huang\affmark[5], Shan You\affmark[6],\\ Chang Xu\affmark[1]\\
\affaddr{\affmark[1] USYD,}
\affaddr{\affmark[2] CSU,}
\affaddr{\affmark[3] SEU,}
\affaddr{\affmark[4] SJTU,}
\affaddr{\affmark[5] HNUST,}
\affaddr{\affmark[6] SenseTime Research}
}

\begin{document}

\maketitle

\begin{abstract}
Modeling dynamic temporal dependencies is a critical challenge in time series pre-training, which evolve due to distribution shifts and multi-scale patterns. This temporal variability severely impairs the generalization of pre-trained models to downstream tasks. 
Existing frameworks fail to capture the complex interactions of short- and long-term dependencies, making them susceptible to spurious correlations that degrade generalization.
To address these limitations, we propose \textbf{DeCoP}, a \textbf{De}pendency \textbf{Co}ntrolled \textbf{P}re-training framework that explicitly models dynamic, multi-scale dependencies by simulating evolving inter-patch dependencies. At the input level, DeCoP introduces Instance-wise Patch Normalization (IPN) to mitigate distributional shifts while preserving the unique characteristics of each patch, creating a robust foundation for representation learning. At the latent level, a hierarchical Dependency Controlled Learning (DCL) strategy explicitly models inter-patch dependencies across multiple temporal scales, with an Instance-level Contrastive Module (ICM) enhances global generalization by learning instance-discriminative representations from time-invariant positive pairs.
DeCoP achieves state-of-the-art results on ten datasets with lower computing resources, improving MSE by 3\% on ETTh1 over PatchTST using only 37\% of the FLOPs.
\end{abstract}

\section{Introduction}

\begin{figure}[t]
  \centering
   \includegraphics[width=1.0\linewidth]{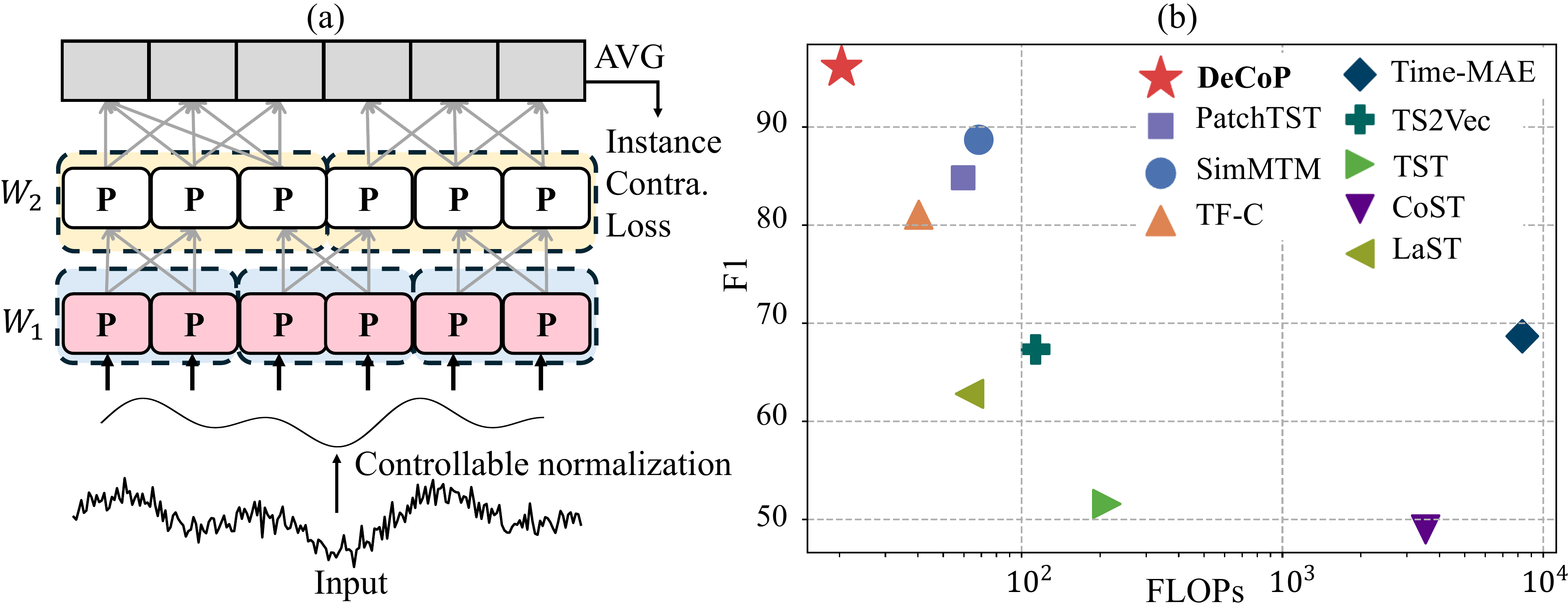}

   \caption{(a) Our dependency-controlled pretraining (DeCoP) framework. DeCoP introduces controlled normalization and instance-level contrastive loss $\mathcal{L}_\text{cl}$ to facilitate dynamic dependency modeling, enabling improved generalization with lower computational overhead. (b) DeCoP consistently outperforms state-of-the-art pretraining frameworks on classification tasks across multiple datasets with lower FLOPs (Floating Point Operations).}
   \label{figure1}

\end{figure}

\begin{figure*}
  \centering
   \includegraphics[width=1\linewidth]{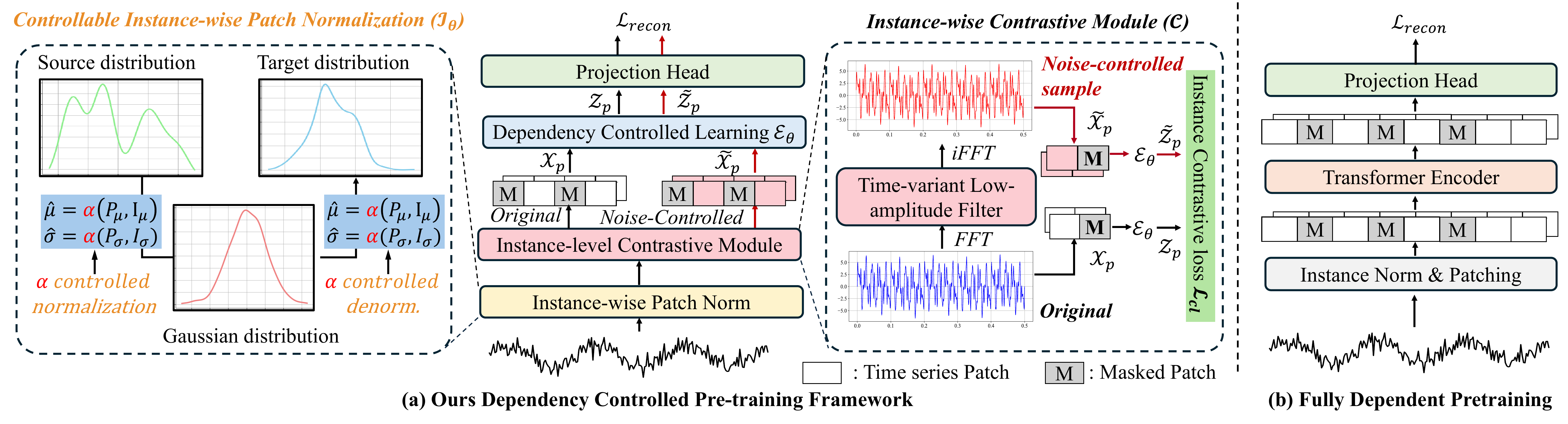}
   \caption{(a) The proposed DeCoP framework addresses dynamic and multi-scale temporal dependencies through a controllable pipeline. IPN incorporates patch-level statistics to stabilize distribution while preserving local semantics. ICM generate time-invariant positive pairs and a contrastive loss to support the hierarchical DCL for multi-scale modeling. DeCoP is optimized with reconstruction loss \( \mathcal{L}_{\text{recon}} \) and contrastive loss \( \mathcal{L}_{\text{cl}} \). 
  (b) In contrast, traditional frameworks neglect finer-scale statistics and fail to capture the complex interactions of multiscale dependencies. 
}
   \label{figure2}

\end{figure*}

Time series analysis plays a critical role in applications like weather forecasting \cite{wu2022timesnet,liu2022non}, fault detection \cite{deng2021graph,zhang2022self}, and sales prediction \cite{wu2023,ekambaram2020attention}. With abundant unlabeled time series data across domains, pre-training approaches for representation learning without extensive annotation are increasingly popular. Recent research has focused on pretrained models to address tasks like forecasting and classification in a general-purpose backbone \cite{goswami2024moment, jin2023timellm, liu2024timer, woo2024unified, rasul2023lag}.

Despite these efforts, time series data pose fundamental challenges for self-supervised learning. First, the non-stationary nature of time series induces temporal distribution shifts, causing the underlying dependency patterns between patches to evolve over time. 
Second, time series inherently exhibit multi-scale temporal structures, encompassing both short-term fluctuations and long-term trends. Models that operate at a single scale consequently fail to capture these rich, hierarchical dependencies. These challenges highlight the need for a framework that can perform controllable, multi-scale dependency modeling in the presence of distribution shifts, enabling representations that capture both fine-grained local semantics and broader contextual patterns.

However, existing time series pre-training frameworks such as Masked Time series Modeling (MTM) \cite{ nie2022time, dong2024simmtm, lee2023learning} predominantly rely on single-scale architectures such as Transformers \cite{vaswani2017attention}. This architectural choice limits their ability to capture multi-scale temporal dependencies and renders them insensitive to relative positional information. Such limitations can lead to spurious correlations, particularly in long-term dependency modeling, and result in entangled representations that blur local variations and dilute global consistency. 
Moreover, to mitigate distribution shifts, these approaches often operate at the instance level and apply uniform normalization statistics across all patches, ignoring their distinct local structures \cite{kim2021reversible}. This coarse-grained normalization results in over-smoothing, suppressing informative temporal patterns such as peaks and abrupt transitions at the patch level.

In this paper, we propose \textbf{DeCoP}, a \textbf{De}pendency \textbf{Co}ntrolled \textbf{P}re-training MTM framework that explicitly models dynamic, multi-scale dependencies across time.
This controllable learning framework enhances the generalization of time series pretrained models while requiring significantly lower computational cost (FLOPs) in \Cref{figure1}. Specifically, at the \textit{input} level, DeCoP applies Instance-wise Patch Normalization (IPN), which incorporate patch-level and instance-level statistics. This enables the model to preserve local semantic variation while stabilizing distribution shifts across time, establishing a more reliable basis for subsequent dependency modeling.

At the \textit{latent representation} level, DeCoP introduces a hierarchical Dependency Controlled Learning (DCL) method to model inter-patch dependencies by dynamically adjusting the temporal receptive field, capturing both short-term and long-range patterns.
Concurrently, we introduce an Instance-level Contrastive Module (ICM) that operates on the representations generated by DCL, promoting global alignment for time-invariant positive pairs to improve performance on high-level downstream tasks such as classification. Extensive experiments demonstrate that DeCoP achieves state-of-the-art performance across ten benchmark datasets. The main contributions of this work are as follows:
\begin{itemize}
\item We propose \textbf{DeCoP}, an efficient dependency controlled pre-training framework that enhances time series representation by explicitly modeling dynamic temporal dependencies under distribution shifts.
\item We introduce Instance-wise Patch Normalization, which integrates patch-level statistical information into normalization. This mitigates distributional shifts and preserves local semantic features captured by patch information, providing a stable foundation for modeling dynamic temporal dependencies.
\item We develop a hierarchical Dependency Controlled Learning strategy that adaptively captures both short- and long-term dependencies across temporal scales, with a Instance-level Contrastive Module aligning high-level semantic information between time-invariant positive sample pairs, enhancing global semantic learning for high-level downstream tasks.
\item DeCoP outperforms existing pretrained models on ten datasets with significantly lower FLOPs, achieving 3\% lower MSE than PatchTST on ETTh1 using only 37\% of the FLOPs.
\end{itemize}

\begin{figure*}
  \centering
   \includegraphics[width=1\linewidth]{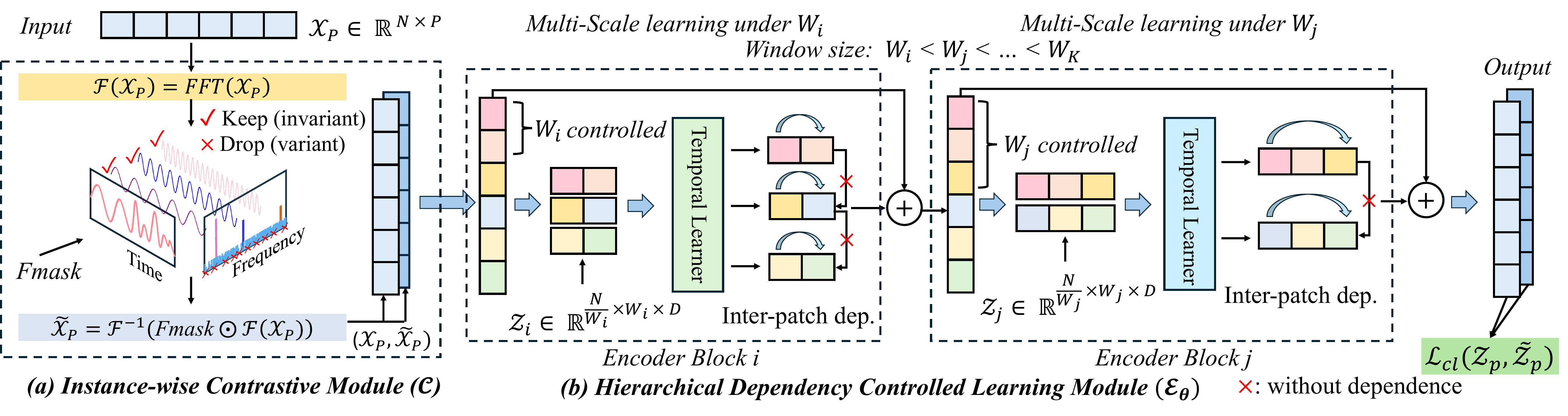}
     \caption{
    (a) The ICM filters time-variant low-amplitude components in the frequency domain to generate denoised positive samples $(\mathcal{X}_p, \tilde{\mathcal{X}_p})$, which serve as stable semantic patterns to facilitate global dependency modeling. 
    (b) The DCL module performs controlled dependency learning across multiple temporal scales, enabling adaptive modeling of inter-patch dependencies under varying temporal dynamics. Inter-patch dep.: inter-patch dependency.
    }
   \label{filter_encoder}
   % \vspace{-3mm}
 
\end{figure*}

\section{Method}

\textbf{Problem Setting}. Given a univariate time series $ x \in \mathbb{R}^{L}$ with look-back length $L$, the output is masked time series patch $\hat{x}$ and the model is optimized by reconstructing the randomly masked patches by $ MSE =  \|x - \hat{x}\|_2^2 $. 

Most existing time series pretraining frameworks follow the transformer structure in Natural Language Processing \cite{devlin2018bert} as shown in \Cref{figure2}b. 
However, this framework overlooks the inherent multi-scale characteristics of time series data and tends to capture spurious dependency correlation due to the dynamical modeling challenge. 
To overcome this, we present \textbf{DeCoP}, a Dependency-Controlled Pre-training framework that improves self-supervised time series representation by modeling dynamic and non-uniform temporal dependencies (\Cref{figure2}a). DeCoP integrates modules at both the input and latent levels. At the input level, instance-wise Patch Normalization stabilizes representations by combining local and instance-level statistics. At the latent level, a hierarchical Dependency Controlled Learning method adaptively models inter-patch dependencies across multiple temporal scales, and an Instance-level Contrastive module generates time-invariant positive samples to enhance global representation modeling. Together, these components enable robust temporal modeling under varying dependencies.

\subsection{Instance-wise Patch Normalization (IPN) for Stabilizing Input Distributions}
\label{sec:AIN}
At the input layer, to address the challenges of distribution shift in time series data for time series dependency modeling, we propose IPN in \Cref{figure2}a. By integrating patch-level variation with global instance-level distribution information, IPN preserves local semantic features, which are critical for capturing short-term patterns and providing a stable foundation for controlled dependency modeling. 
Specifically, given a univariate time series $ x \in \mathbb{R}^{L}$, we divide it into patches \(\mathcal{X}_p =\{x_1, x_2, ..., x_N\}\) with patch size \(P\) and stride \(S\).  The total number of patches $N$ is given by:
\begin{align}
 N = \left\lfloor \frac{L - P}{S} \right\rfloor + 2.
\end{align} 
Each patch $x_n \in \mathbb{R}^P$ is an independent unit that captures localized temporal patterns. To quantify the fine-grained variations within each patch, we first compute the patch-wise mean, $E_P[x_n]$, by averaging over its $P$ time points:
\begin{align}
E_{P} \left[ x_n \right] = \frac{1}{P} \sum_{i=1}^{P} x_{n,i}.
\end{align}
Subsequently, we compute the patch-wise variance, $Var_P[x_n]$, which measures the dispersion of these points around the mean:
\begin{align}
Var_{P} \left[ x_n \right] = \frac{1}{P} \sum_{i=1}^{P} \left( x_{n,i} - E_{p} \left[ x_{n} \right]\right)^2.
\end{align}
To exploit instance statistics, we leverage distribution information at the instance scale to incorporate global distribution information by calculating mean and variance of $x$: 
\begin{align}
E_{I} \left[ x \right] &= \frac{1}{L} \sum_{j=1}^{L} x_{j},
Var_{I}\left[ x \right] = \frac{1}{L} \sum_{j=1}^{L} \left( x_{j} - E_I \left[ x \right] \right)^2.
\end{align}
where $j$ is the relative index of time series $x$. For each time series, the mean and variance are calculated along the $L$ dimension.
After obtaining instance and patch-wise distribution information, a learnable parameter $\alpha \in \mathbb{R}$ is introduced to balance local and global information, controlling their influence as follows:
\begin{align}
E \left[ x_n \right] &= (1-\alpha) \times E_I + \alpha \times E_P,
\end{align}
where $E_I$ and $E_P$ are the global mean and local mean of $x_n$, respectively. The calculation of final variations is given by:
\begin{align}
Var\left[ x_n \right] &= (1-\alpha) \times Var_I + \alpha \times Var_P.
\end{align}
Finally, each patch $x_n$ is transformed into $\Tilde{{x}}_n$ using the computed mean and variance for normalization:
\begin{align}
\Tilde{{x}}_n = \frac{x_n - E \left[ x_n \right]}{\sqrt{{Var}\left[ x_n \right] + \epsilon}}.
\end{align}
This results in a normalized sequence $\mathcal{X}_p = \{\Tilde{{x}}_1, \Tilde{{x}}_2, ...,\allowbreak \Tilde{{x}}_n,..., \allowbreak\Tilde{{x}}_N\}$. During the pretraining stage stage, we reconstruct the normalized time series $\mathcal{X}_p$. During the finetuning stage, we return the mean and variance back through denormalization. This process establishes a stable foundation for dependency modeling, effectively overcoming the distribution shift problems.

\subsection{Hierarchical Dependency Controlled Learning (DCL) for Dynamic Dependencies}
\label{DCL}
To address the dynamic and multiscale temporal dependencies in time series, we propose a DCL module that adaptively controls the receptive field at the latent representation level (\Cref{filter_encoder}b). Temporal dependencies vary in temporal range, which requires the model to capture both short- and long-range patterns. Our multi-scale design enables the model to flexibly adjust the dependency range and effectively capture multi-scale patterns across varying temporal structures.

Specifically, given the denoised positive pair $(\mathcal{X}_p, \mathcal{\Tilde{X}}_p)$ (discussed in next section), we omit the modeling of $\mathcal{\Tilde{X}}_p$ as both inputs share the same encoder backbone. The input patches $\mathcal{X}_p \in \mathbb{R}^{N \times P}$ are first projected into a latent space via a linear transformation:
\begin{align}
    \mathcal{Z}_p &= \mathcal{X}_p W_P + Bias,\quad
    \mathcal{Z}_p = \mathcal{Z}_p + W_{pos},
\end{align}
where $W_P \in \mathbb{R}^{P \times D} $, $Bias \in \mathbb{R}^{D} $ and $D$ is the model dimension. 
To retain the temporal structure, we incorporate a learnable relative positional encoding $ W_{pos} \in \mathbb{R } ^ {N \times D}$  into the latent representation $\mathcal{Z}_p$.
To capture dependencies across different temporal scales, we define a set of window sizes $ \left\{ W_k \right\}_{k=1}^K$ for hierarchical modeling.  
For each window size, we reshape the $\mathcal{Z}_p$ into a windowed form and flatten the local window dimensions.
\begin{align}
\mathcal{Z}_r &= Flatten(Reshape(Padding(\mathcal{Z}_p))).
\end{align}
Specifically, $Z_p$ is first padded and reshaped into the defined window size $W_k$, converting from $\mathbb{R}^{N \times D}$ to $\mathbb{R}^{N/W_k, W_k, D}$, and then flatten the last two dimensions into $\mathbb{R}^{N/W_k, W_k \times D}$. This enables the encoder to operate over local windows of varying sizes. Then, the resulting representation within each window is passed to a temporal learner $\mathcal{E}_{\text{enc}}$:
\begin{align}
\mathcal{Z}_{e} &= ReshapeBack(\mathcal{E}_{enc}(\mathcal{Z}_r)), 
\end{align}
where $\mathcal{E}_{enc}$ is based on a Linear or MLP structure in our experiment. After that, we reshape the input $\mathcal{Z}_{e}$ to its original size $\mathbb{R}^{N \times D}$ and leverage residual connections to ensure training stability as follows:
\begin{align}
\mathcal{Z}_k &= \mathcal{Z}_{e} + Dropout(\mathcal{Z}_p).
\end{align}
The window size $W_k$ increases across encoder blocks, enabling a global receptive view through careful design. This scale-aware structure empowers DCL to adaptively learn temporal dependencies from both local and global perspectives, making it suitable for handling varying and evolving temporal structures in time series.

\begin{figure*}
  \centering
   \includegraphics[width=1\linewidth]{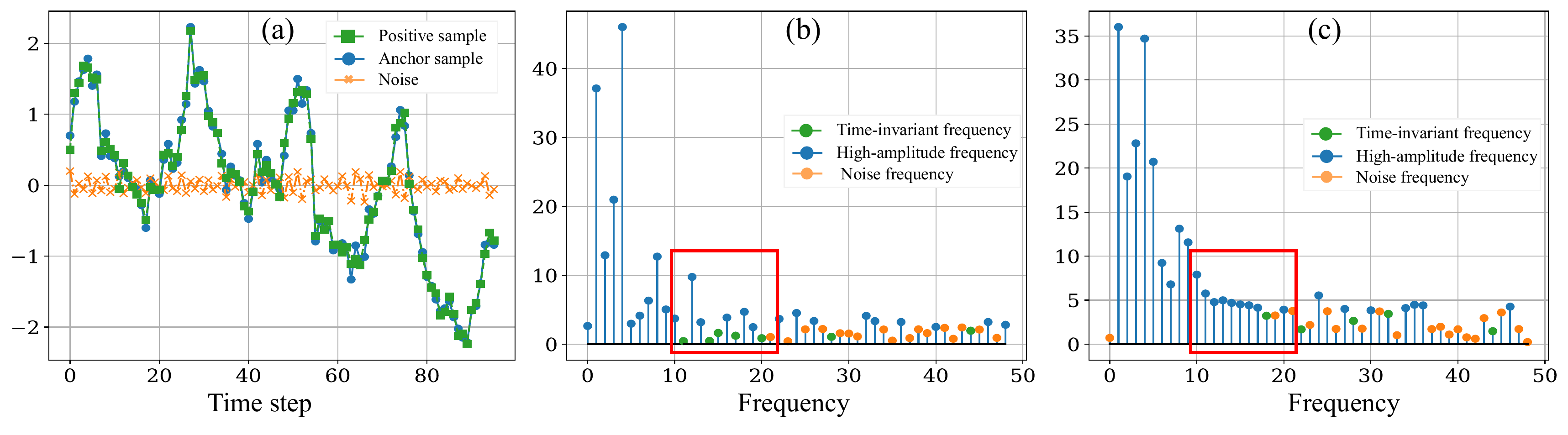}
\caption{ The CFD module filters time-variant noise while preserving meaningful time-invariant frequencies. 
(a) Visualization of the anchor sample, filtered noise, and generated positive sample on the last channel of the ETTh1 dataset (sequence length = 100). The filtered noise (orange) resembles white noise with zero mean, consistent with the central limit theorem. 
(b) Amplitude spectrum of (a), where green dots denote time-invariant frequencies retained by CFD, and orange dots indicate time-variant frequencies removed by the $Fmask$. 
(c) Although the green-highlighted frequencies in (b) show lower amplitude, they remain prominent across (c), confirming their time-invariant nature and importance for capturing stable patterns.
}
   \label{figure7}
\end{figure*}

\begin{algorithm}[t]
\caption{DeCoP Framework}
\KwIn{$\mathcal{X}_p$: Patched time series;\\
$\mathcal{I}_\theta$: Instance-wise patch normalization module;\\
$\mathcal{C}$: Controllable frequency decomposition module;\\
$\mathcal{E}_\theta$: Hierarchical dependency-controlled learning module with temporal learner $\mathcal{E}_{enc}$;\\
$\mathcal{L}_{cl}$: Contrastive loss for global temporal semantics;\\
$\mathcal{L}_{recon}$: Reconstruction loss for recovering masked patches;\\
% $\mathcal{H}$: Prediction head for reconstructing masked patches;\\
$\gamma$: Weighting factor to balance $\mathcal{L}_{cl}$ and $\mathcal{L}_{recon}$;
}
\For{epoch in range(Epochs)}{
    Normalize $\mathcal{X}_p$ via $\mathcal{I}_\theta(\mathcal{X}_p)$;\\
    Generate positive pairs $(\mathcal{X}_p, \tilde{\mathcal{X}}_p)$ via $\mathcal{C}(\mathcal{X})$;\\
    Randomly mask pairs $(\mathcal{X}_p, \tilde{\mathcal{X}}_p)$ with $\text{FMask}$;\\
    Calculate latent representation $(\mathcal{Z}_k, \tilde{\mathcal{Z}_k}) = \mathcal{E}_\theta(\mathcal{X}_p, \tilde{\mathcal{X}}_p)$, where $\mathcal{E}_\theta$ includes $(\mathcal{Z}_e, \tilde{\mathcal{Z}_e}) = \mathcal{E}_{enc} (\mathcal{X}_p, \tilde{\mathcal{X}}_p)$;\\
    Calculate $\mathcal{L}_{cl}(\mathcal{Z}_e, \tilde{\mathcal{Z}}_e) = - \sum \log \mathcal{P}(Z_e, \tilde{Z}_e)$;\\
    Reconstruct masked patches ($\hat{\mathcal{X}}_p$, $\hat{ \tilde{\mathcal{X}}}_p$)\\
    % via $\mathcal{H}(\mathcal{Z}_k, \tilde{\mathcal{Z}_k})$;\\
    Compute loss $\mathcal{L} = \mathcal{L}_{recon}(\mathcal{X}_p, \hat{\mathcal{X}}_p) + \gamma \mathcal{L}_{cl}$;\\
    Optimize parameter $\theta$ with Adam optimizer;
}
\textbf{end}
\label{algorithm1}
\end{algorithm}

\subsection{Instance-level Contrastive Module (ICM) for Global Semantics}
\label{sec:FTC}
We introduce ICM to enhance DCL for global semantic modeling, as MTM paradigm may be suboptimal for high-level downstream tasks such as classification compared to CL paradigm (\Cref{filter_encoder}a and \Cref{figure7}). To construct positive sample pairs for DCL's high-level semantic modeling, we incorporate a controllable time-variant low-amplitude filtering to generate noise-less views. 
Specifically, we first apply a Discrete Fourier Transform $\mathcal{F}$ to the batched input $\mathcal{X}_b = [\mathcal{X}^0_p, \mathcal{X}^1_p, ..., \allowbreak \mathcal{X}^M_p] \in \mathbb{R}^{B, M, N}$,  where $B$ denotes batch size, $M$ denotes channel size, resulting in the corresponding spectrum: $S_b = \mathcal{F}(X_b)$.
where the generated $\mathcal{S}_b$ has shape $[B, M, \lfloor \frac{L}{2} \rfloor]$.  To identify stable global frequency components, $\mathcal{S}_{\text{mean}}$ across the batch and channel dimensions, we compute the average amplitude:
\begin{align}
\mathcal{S}_{\text{mean}} = \text{AVG}(\text{AVG}(\text{AMP}(\mathcal{S}_b), \text{dim}=0), \text{dim}=0),
\end{align}
where $\text{AMP}$ denotes the calculation of amplitude, and $\text{AVG}$ represents the average operation. Then, we apply a $\text{top}K$ controlled filter operation to keep the global salient frequency with high amplitude, yielding the time- and channel-invariant set $\mathcal{S}_{\text{invar}}$.
\begin{align}
\mathcal{S}_{\text{invar}} = \text{Filter}\left( \text{top}K, \mathcal{S_{\text{mean}}} \right),
\end{align}
where $\text{top}K = \beta \times \lfloor L /2\rfloor$, and $\beta$ is a hyperparameter that controls the filtering intensity.
We then introduce a $\text{top}M$-controlled low-amplitude filter to obtain instance-wise low-amplitude frequencies.
\begin{align}
\mathcal{S}_{\text{var}} = \text{Filter}\left( \text{top}M, \text{AMP}(\mathcal{S}_b)\right),
\end{align}
where $\text{top}M = (1-\beta) \times \lfloor L /2\rfloor$. To prevent information loss, the final time-variant filtered $\mathcal{S_{\text{var}}}$ is defined as:
\begin{align}
\mathcal{S_{\text{var}}} = \mathcal{S_{\text{var}}} - \mathcal{S_{\text{var}}} \cap \mathcal{S_{\text{invar}}}.
\end{align}
We construct a binary frequency mask $\text{FMask} \in \{0, 1\}^{B \times M \times \lfloor L/2 \rfloor}$ to suppress time-variant components. Specifically, each element $\text{FMask}_{b,m,k}$ is set to 0 if $i \in \mathcal{S}_{\text{var}}$, and 1 otherwise, where $k \in [0, \lfloor L/2 \rfloor)$, $b \in [1, B]$, and $m \in [1, M]$. $\text{FMask}$ is applied to obtain a filtered spectrum. The denoised signal $\tilde{X}_b$ is then recovered by applying the inverse Fourier transform $\mathcal{F}^{-1}$:
\begin{align}
\tilde{X}_b = \mathcal{F}^{-1}(\text{FMask} \odot \mathcal{S}_b),
\end{align}
where $\odot$ denotes the Hadamard product. Once the batched positive pairs $(\mathcal{X}_b, \mathcal{\Tilde{X}}_b)$ are obtained, where $\mathcal{\Tilde{X}}_b$ preserves reliable temporal structures, we apply random masking $m$ to each pair $(\mathcal{X}_p, \mathcal{\Tilde{X}}_p)$ from the batch for subsequent masked modeling. For each positive pair, the temporal learner $\mathcal{E}_{{enc}}$ extracts their latent representations as:
\begin{align}
\mathcal{Z}_e = \mathcal{E}_{{enc}}(\mathcal{X}_p),\quad \mathcal{\Tilde{Z}}_e = \mathcal{E}_{{enc}}(\mathcal{\Tilde{X}}_p).
\end{align}
We further define a similarity-based contrastive loss function $\mathcal{L}_{cl}$ for the final DCL block as follows:
\begin{align}
    \mathcal{L}_{cl} &= 1 - \frac{1}{BM} \sum_{i=1}^B \sum_{m=1}^M \text{AVG}(\mathcal{Z}_e) \cdot \text{AVG}(\mathcal{\Tilde{Z}}_e),
\end{align}
where $\cdot$ denotes the dot product.
By minimizing $\mathcal{L}_{cl}$, the model encourages $ \mathcal{Z}_e $ to approximate its denoised counterpart $ \tilde{\mathcal{Z}}_e $, enhancing global instance-level representation learning and improving generalization to downstream tasks.

\subsection{Loss Design}
Based on MTM framework, we reconstruct masked patched rely on unmasked patch. A pretraining head is used to predict the masked patches as follows:
\begin{align}
    \Hat{\mathcal{X}_p} = \mathcal{Z}_k W + Bias,
\end{align}
where $W \in \mathbb{R}^{D \times P} $, $Bias \in \mathbb{R}^{P} $  and $\mathcal{Z}_k \in \mathbb{R}^{N \times D} $. Here, we also adopt MSE as the reconstruction loss:
\begin{align}
\mathcal{L}_{recon} = \sum_{i=1}^B \sum_{m=1}^M \sum_{n=1}^N \left\lVert m \odot (\mathcal{X}_p^{(i, m, n)} - \hat{\mathcal{X}}_p^{(i, m, n)})\right\rVert_2^2,
\end{align}
where $m$ is a binary mask indicating whether a patch is masked and $\odot$ denotes the Hadamard product. The final loss combines the reconstruction loss with the contrastive loss previously introduced:
\begin{align}
\mathcal{L} = \mathcal{L}_{recon} + \gamma \times \mathcal{L}_{cl},
\end{align}
where the hyperparameter $\gamma$ is the weight of $\mathcal{L}_{cl}$. This combined loss enhance temporal consistency under temporal noise and non-uniform dependencies, encouraging the model to learn representations that preserve global patterns. 

\begin{table*}[ht]
    \centering
    \resizebox{\textwidth}{!}{%
    \begin{tabular}{lcccccccccccccc}
    \toprule
    \multirow{2}{*}{Model} & \multicolumn{2}{c}{ETTh1} & \multicolumn{2}{c}{ETTh2} & \multicolumn{2}{c}{ETTm1}& \multicolumn{2}{c}{ETTm2} & \multicolumn{2}{c}{Weather} & \multicolumn{2}{c}{Electricity} & \multicolumn{2}{c}{\textbf{Average}} \\
    \cmidrule(lr){2-3} \cmidrule(lr){4-5} \cmidrule(lr){6-7} \cmidrule(lr){8-9} \cmidrule(lr){10-11} \cmidrule(lr){12-13} \cmidrule(lr){14-15}
    & MSE & MAE & MSE & MAE & MSE & MAE & MSE & MAE & MSE & MAE & MSE & MAE & MSE & MAE \\
    \midrule
    Informer & 1.033	&0.799 &3.303	&1.439& 0.872&	0.691 &1.305	&0.797 &0.568	&0.522 &0.329	&0.415 &1.235&	0.777\\
    Autoformer  &0.473	&0.477 &0.422	&0.443& 0.515	&0.493& 0.310	&0.357& 0.335&	0.379& 0.214&	0.327 &0.378	&0.412\\
    Fedformer  & 0.428 &0.454 & 0.388	&0.434 &0.382	&0.422 & 0.292	&0.343 &0.310&	0.357& 0.207&	0.321 &0.335&	0.389 \\
    iTransformer & 0.479 &0.477 & 0.387	&0.418 &0.371	&0.400& 0.272	&0.333 &0.246&	0.278& 0.161&	0.256 &0.319&	0.360 \\
    DLinear  &  0.433&	0.446& 0.477&	0.469& 0.360&	0.384 &0.283	&0.346 &0.247&	0.311& 0.162	&0.260& 0.327&	0.369\\
    TimeMixer  & 0.432&	0.446 &0.375&	0.413 &	0.395	&0.389&0.262&	0.322& 0.228&	0.269 & 0.165& 0.261& 0.309& 0.353 \\
    CycleNet & 0.430&	0.440 &0.367	&0.406 & 0.368	&0.395 &0.267 &0.325& \underline{0.224}	&\underline{0.265} &0.158	&0.252& 0.302&	0.347\\
    PatchTST  & 0.430&	0.445  &0.355&	0.394  &0.346	&0.383 &0.257	&0.318 &0.225	&0.261	&\textbf{0.157}&	\underline{0.252}&0.295	&0.342\\
    SimMTM  &  0.404	&0.428 & 0.348&	0.391 &0.362	&0.393 &0.269	&0.327& 0.227&	0.268& 0.162&	0.256 &0.295	&0.344\\
    \midrule
    \rowcolor{gray!15} \textbf{$\text{DeCoP}_{\text{Linear}}$}&  \textbf{0.401}&	\textbf{0.421}&	\textbf{0.333}&	\textbf{0.382}&	\underline{0.361}&	\underline{0.379} &\underline{0.255}	&\underline{0.313}&	0.242&	0.279	&0.165&	0.258& \underline{0.293}& \underline{0.339}	\\
    \rowcolor{gray!15} \textbf{$\text{DeCoP}_{\text{MLP}}$} &  \underline{0.408} &	\underline{0.424}& \underline{0.341} &\underline{0.388}& \textbf{0.342}	&\textbf{0.376} &\textbf{0.249}	&\textbf{0.311}& \textbf{0.223}&	\textbf{0.259}& \textbf{0.157}&	\textbf{0.251}& \textbf{0.287}& \textbf{0.335}\\
    \bottomrule
    \end{tabular}
    }
    \caption{Forecasting results for predicting $F$ future time points based on the past 512 points in an in-domain setting. Results are averaged over $F \in \{96, 192, 336, 720\}$, with lower MSE and MAE indicating better performance. The best and second-best results are highlighted in \textbf{bold} and \underline{underlined}, respectively.}
    \vspace{-3mm}
    \label{table2}
\end{table*}

\begin{table*}[ht]
    \centering
    \resizebox{1.0\textwidth}{!}{
    \begin{tabular}{cc|>{\columncolor{gray!15}}c>{\columncolor{gray!15}}c|>{\columncolor{gray!15}}c>{\columncolor{gray!15}}c|cc|cc|cc|cc|cc|cc}
        \toprule
        \multicolumn{2}{c|}{Scenarios}& \multicolumn{2}{c|}{\cellcolor{gray!15}\textbf{$\text{DeCoP}_{\text{Linear}}$}} & \multicolumn{2}{c|}{\cellcolor{gray!15}\textbf{{$\text{DeCoP}_{\text{MLP}}$}}} & \multicolumn{2}{c|}{PatchTST} & \multicolumn{2}{c|}{SimMTM} & \multicolumn{2}{c|}{TimeMAE} &\multicolumn{2}{c|}{CoST} &\multicolumn{2}{c|}{TST}&\multicolumn{2}{c}{TF-C} \\
        \cmidrule(lr){1-2} \cmidrule(lr){3-4} \cmidrule(lr){5-6} \cmidrule(lr){7-8} \cmidrule(lr){9-10} \cmidrule(lr){11-12} \cmidrule(lr){13-14} \cmidrule(lr){15-16} \cmidrule(lr){17-18}
        Source & Target & MSE & MAE & MSE & MAE & MSE & MAE & MSE & MAE & MSE & MAE & MSE & MAE & MSE & MAE & MSE & MAE\\
        \midrule
         ETTh2 & ETTh1 &  \textbf{0.403} &  \textbf{0.422} &\underline{0.409} &\underline{0.426}&0.423	& 0.443 & 0.415 &0.43 & 0.466	&0.456 & 0.428 & 0.433	& 0.469	& 0.459 & 0.635	&0.634\\
        ETTm2 & ETTm1 & 0.359 &\underline{0.379} &  \textbf{0.342} &  \textbf{0.376}&	\underline{0.348} &0.382 &	0.351 &	0.383 &0.390 &0.410 & 0.385&	0.412&	0.382&	0.402 &0.758&	0.669\\
         ETTm2 & ETTh1 &  \textbf{0.404}&  \textbf{0.423}&\underline{0.412}	& \underline{0.426}	&0.433	& 0.447 &0.428 & 0.441  &0.464	&0.456 & 0.598 &	0.548 &	0.453 &	0.45 &1.091	&0.814\\
         ETTh2 & ETTm1  &\underline{0.360} &\underline{0.379} & \textbf{0.343} &  \textbf{0.377}&	0.363 &	0.387 &	0.365 &	0.384  &0.383	&0.402 & 0.363	&0.387&	0.391	&0.409 &0.750 &0.654\\
          ETTm1 & ETTh1  &  \textbf{0.405} &  \textbf{0.423}&\underline{0.416}&\underline{0.427} &	0.447&	0.451&	0.422&	0.430  &0.495&	0.469 & 0.62&	0.541&	0.475&	0.463 &0.700&	0.702\\
          ETTh1 & ETTm1 & 0.361	&0.379	& \textbf{0.346}	& \textbf{0.379}		&\underline{0.348}	&\underline{0.381}	&0.346	&0.384  &0.360&	0.390 & 0.37	&0.393	&0.373	&0.393 &0.746	&0.652\\
          Weather & ETTh1  &  \textbf{0.405}	& \textbf{0.422}	&\underline{0.411}	&\underline{0.426}		&0.437	&0.448 &0.456	&0.467 & 0.518&	0.487&	0.465&	0.456
          &0.462	&0.464 &-	&- \\
          Weather & ETTm1 &0.359& \underline{0.379}	& \textbf{0.345}	& \textbf{0.376}	&0.348	&0.383	&0.358	&0.388 &0.411	&0.423 & 0.382&	0.403&	0.368&	0.392 &-	&- \\  
          \midrule
           \multicolumn{2}{c|}{\textbf{Average}}& \underline{0.382}&\textbf{0.401}&\textbf{0.378}&\underline{0.402}&0.393&0.415&0.393&0.413 
          & 0.429&	0.434 &0.458	&0.451	&0.422	&0.428 &0.780	&0.693\\
        \bottomrule
    \end{tabular}
    }
    \caption{Transfer learning setting of forecasting the future $F$ time points. All results are averaged from 4 different choices of $F \in \{96, 192, 336, 720\}$. The best and second-best results are highlighted in \textbf{bold} and \underline{underlined}, respectively.}
    % \vspace{-3mm}
    \label{table3}
\end{table*}

\section{Experiments}
\label{sec:Experiment}
\textbf{Experiment Setting}. We conduct experiments on forecasting and classification tasks, following the protocols in \cite{nie2022time} and \cite{dong2024simmtm}. Fine-tuning performance is evaluated under in-domain and cross-domain settings. MSE and MAE are used as metrics for forecasting, while Accuracy, Precision, Recall, and F1-score assess classification performance. For forecasting, six real-world datasets are employed, including four ETT datasets \cite{zhou2021informer} (ETTh1, ETTh2, ETTm1, ETTm2), Weather \cite{weatherdata}, and Electricity \cite{ecldata}. The ETT datasets were collected from two distinct electric transformers over a two-year period. The Weather dataset comprises 21 meteorological indicators recorded every 10 minutes in Germany. The Electricity dataset contains hourly electricity consumption records of 321 customers.

For classification, we adopt four real-world datasets: SleepEEG \cite{kemp2000analysis-sleepeeg}, Epilepsy \cite{EPILEPSY}, FD-B \cite{FD-B}, and EMG \cite{EMG}. SleepEEG includes 153 full-night EEG recordings across five sleep stages. Epilepsy contains single-channel EEG from 500 subjects with binary seizure labels. FD-B involves motor bearing signals classified into three fault types: undamaged, inner-damaged, and outer-damaged. EMG records muscle responses from three patients with neuropathy or myopathy, each representing a class. The details of datasets are provided in supplementary material.

\begin{table*}[t]
    \centering
    \resizebox{1\textwidth}{!}{
    \setlength{\tabcolsep}{2mm}
    \begin{tabular}{l|cccc|cccc|cccc|cccc}
        \toprule
        \multirow{2}{*}{Scenarios} & \multicolumn{4}{c|}{In-domain} & \multicolumn{12}{c}{Cross-domain} \\
        \cmidrule(lr){2-17} 
        & \multicolumn{4}{c|}{Epilepsy $\to$ Epilepsy} & \multicolumn{4}{c|}{SleepEEG $\to$ Epilepsy}& \multicolumn{4}{c|}{SleepEEG $\rightarrow$ FD-B} & \multicolumn{4}{c}{SleepEEG $\rightarrow$ EMG}\\
        \midrule
         Metrics & ACC & P & R & F1 & ACC & P & R & F1 & ACC & P & R &F1  & ACC & P & R & F1  \\
        \midrule
        TS2Vec & 92.17 & 93.84 & 81.19 & 85.71 & 93.95 & 90.59 & 90.39 & 90.45 & 47.9 & 43.39 & 48.42 & 43.89 & 78.54 & 80.4 & 67.85 & 67.66 \\
        LaST& 92.11 & 93.12 & 81.47 & 85.74 & 86.46 & 90.77 & 66.35 & 70.67 & 46.67 & 43.9 & 47.71 & 45.17 & 66.34 & 79.34 & 63.33 & 72.55\\
        TF-C & 93.96 & 94.87 & 85.82 & 89.46 & 94.95 & \textbf{94.56} & 80.08 & 91.49 & 69.38 & 75.59 & 72.02 & 74.87 & 81.71 & 72.65 & 81.59 & 76.83 \\
        TST & 80.21 & 40.11 & 50.00 & 44.51 & 80.21 & 40.11 & 50.00 & 44.51 & 46.4 & 41.58 & 45.5 & 41.34  & 78.34 & 77.11 & 80.3 & 68.89\\
        CoST& 88.07 & 91.58 & 66.05 & 69.11 & 88.40 & 88.20 & 72.34 & 76.88 & 47.06 & 38.79 & 38.42 & 34.79 & 53.65 & 49.07 & 42.1 & 35.27 \\
        Ti-MAE & 90.90 & 93.90 & 77.24 & 78.21 & 89.71 & 72.36 & 67.47 & 68.55 & 60.88 & 66.98 & 68.94 & 66.56 & 69.99 & 70.25 & 63.44 & 70.89 \\
        PatchTST& 89.56&90.39&89.56&80.11&93.27&92.51&85.57&88.48 &  \underline{80.15}&	\underline{82.25} &\underline{85.47} &\underline{83.05} & 90.24&82.96&82.95&82.91 \\ 
        SimMTM  & \underline{94.75} & \textbf{95.6} & \underline{89.93} & \underline{91.41}  & \underline{95.49} & \underline{93.36} & \underline{92.28} & \underline{92.81} & 69.40& 74.18 & 76.41 & 75.11 & \underline{97.56} & \underline{98.33} & \underline{98.04} & \underline{98.14} \\
        \midrule
        \rowcolor{gray!15} \textbf{{$\text{DeCoP}_{\text{MLP}}$}} & \textbf{95.53} & \underline{93.51} & \textbf{92.25} & \textbf{92.86} & \textbf{95.82} & 94.23 & \textbf{92.41} & \textbf{93.28} & \textbf{93.04} & \textbf{94.92} & \textbf{94.90} & \textbf{94.90}& \textbf{100} & \textbf{100} & \textbf{100} & \textbf{100}\\
        \bottomrule
    \end{tabular}
    }
    \caption{In- and cross-domain classification. For in-domain, DeCoP is pretrained and finetuned on Epilepsy. For the cross-domain setting, we pretrain DeCoP on SleepEEG and fine-tune it to multiple target datasets: Epilepsy, FD-B, EMG. P and R denotes precision and recall, respectively. The best and second-best results are highlighted in \textbf{bold} and \underline{underlined}, respectively.}
    \label{tabel4}
\end{table*}

\begin{figure*}[ht]

  \centering
    \includegraphics[width=1\linewidth]{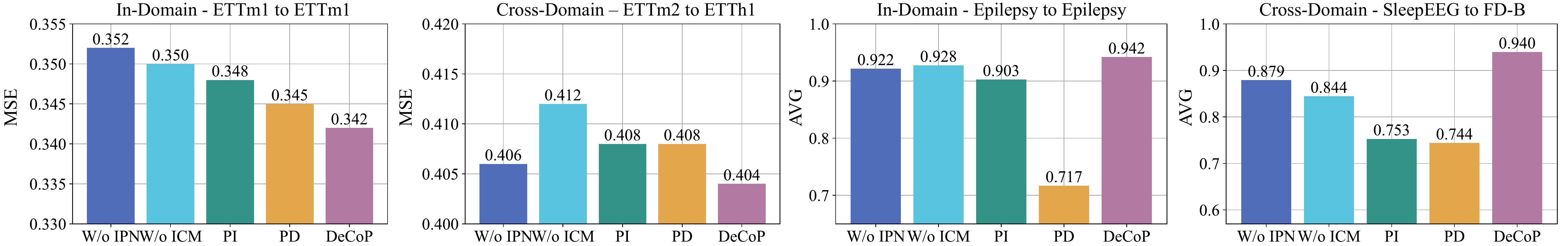}
   \caption{Ablation study of DeCoP, showing the impact of IPN, ICM, DCL, Patch Dependence (PD), and Patch Independence (PI) on time series forecasting (left) and classification (right) tasks in both in-domain and cross-domain settings. For classification, AVG denotes the average of accuracy and F1 score.}
   % \vspace{-3mm}
   \label{figure5}
\end{figure*}

\noindent \textbf{Model Parameters}. By default, all experiments are configured with the following parameters: $k = 2$, top$K = 0.3$, $\alpha_{initial} = 0.01$, $\gamma = 0.1$ and $lr = 1e-4$. For forecasting tasks, both in- and cross-domain experiments share the same configuration, with a patch size and stride of 12. For both in- and cross-domain classification task, the patch size and stride are set to 8. More details of parameters are provided in supplementary material.

\subsection{Time series Forecasting}
\textbf{In-domain Evaluation}. We compare our model with six competitive state-of-the-art baseline methods in time series forecasting, including self-supervised approaches
(SimMTM \cite{dong2024simmtm}, PatchTST \cite{nie2022time}) and supervised approaches (CycleNet \cite{lin2024cyclenet}, TimeMixer \cite{wang2024timemixer}, DLinear \cite{zeng2023transformers}, iTransformer \cite{liu2023itransformer}, Fedformer \cite{zhou2022fedformer}, Autoformer \cite{wu2021autoformer}, Informer \cite{zhou2021informer}). 
In \Cref{table2}, $\text{DeCoP}_{\text{Linear}}$ outperforms the second-best by 1.5\% on ETTh2. For more complex datasets like ETTm2, $\text{DeCoP}_{\text{MLP}}$ achieves the best results, surpassing the PatchTST by 0.8\% in MSE. 

\noindent \textbf{Cross-domain Analysis}. In the cross-domain setting, we compare our framework with six advanced time series pre-training frameworks SimMTM, PatchTST, TF-C \cite{Zhang2022-TF-C}, TST \cite{zerveas2021transformer}, CoST \cite{woo2022cost} and TimeMAE \cite{li2023ti-TiMAE}). In \Cref{table3}, we evaluate multiple scenarios to test effectiveness under cross-domain conditions.
Both in-domain and cross-domain transfer settings, our model consistently achieves lower MSE and MAE than others, especially in ETTm1 $\to$ ETTh1, we outperform PatchTST 4.2\% in MSE, highlighting its effectiveness under distribution shifts. 
% Complete forecasting results are provided in the supplementary material.

\subsection{Time series Classification}
\textbf{In-domain Evaluation}. For in-domain learning, We preform {\allowbreak Epilepsy $\to$ Epilepsy} following \cite{dong2024simmtm}. We adopt MLP as our temporal learner in classification task and compare it with eight competitive state-of-the-art baseline methods, including the contrastive learning based methods: TF-C, LaST, TST, TS2Vec \cite{yue2022ts2vec}, and the masked time series modeling methods: SimMTM, PatchTST, Ti-MAE, CoST \cite{wang2022learning}. In \Cref{tabel4}, Our model outperform second-best SimMTM by 1.45\% in F1, and outperform PatchTST by 12.75\%.

\noindent \textbf{Cross-domain Analysis}. For cross-domain setting, we conduct experiments across in-domain and cross-domain transfer learning SleepEEG → \{Epilepsy, FD-B, EMG\} in \Cref{tabel4}, where the source data differs from the target data in both channels and classes. Notably, on SleepEEG $\to$ FD-B, Our framework surpass the second best by 12.89\% and 11.85\% in accuracy and F1, respectively. These results highlight DeCoP’s superior robustness under both domain and label shifts. 
% Complete classification results are provided in supplementary material.

\subsection{Model Analysis}
\textbf{Ablations}. 
We conduct ablation studies on both forecasting and classification tasks under in-domain and cross-domain settings to evaluate the contributions of IPN, ICM, and DCL. For DCL, we compare two alternative configurations: patch-independent (PI) and patch-dependent (PD). In \Cref{figure5}, replacing DCL with PI or PD leads to average performance drops of 3.94\% and 22.51\% on in-domain classification, respectively. The performance gap becomes larger in the cross-domain setting, with declines of 18.69\% (PI) and 19.57\% (PD), highlighting the importance of dynamical dependency modeling. In forecasting, removing IPN and ICM results in MSE increases of 0.5\% and 0.4\% under in-domain and cross-domain settings, respectively. 

Notably, ICM contributes to more stable gains across both forecasting and classification: its removal increases forecasting error by up to 0.8\% in the ETTm2 $\to$ ETTh1 task and reduces the average of F1 and accuracy by 9.53\% in the SleepEEG $\to$ FD-B scenario. These results confirm the effectiveness of ICM in enhancing generalization to downstream tasks, particularly for high-level classification tasks. 

\textbf{Better Results with Reduced FLOPs}. We compute the FLOPs and parameters of DeCoP compared to two SOTA pre-training frameworks in two datasets in \Cref{table_flops}. In both pre-training and fine-tuning stages, DeCoP achieves the lowest MSE 0.401, outperforming PatchTST by 30\% on the ETTh1 dataset while using only 37\% of the FLOPs. 

\begin{table}[t]
\centering
\resizebox{0.48\textwidth}{!}{
\begin{tabular}{ccccccc}
\toprule
\multirow{2}{*}{Dataset} & \multirow{2}{*}{Models}  & \multicolumn{2}{c}{Pretrain} & \multicolumn{2}{c}{Finetune} & \multirow{2}{*}{MSE} \\  \cmidrule(lr){3-4} \cmidrule(lr){5-6}
&  & FLOPs         & Params        & FLOPs         & Params        \\ \midrule              
\multirow{3}{*} {ETTh1} & PatchTST & 175M   & 0.598M   & 130M   & \textbf{2M} & 0.430  \\ 
& SimMTM & 4269M     & 143M     & 100M &  6M     & 0.404   \\ 
 & \cellcolor{gray!15} \textbf{DeCoP}    &\cellcolor{gray!15} \textbf{72M}    & \cellcolor{gray!15}\textbf{0.479M}   &\cellcolor{gray!15} \textbf{49M}   & \cellcolor{gray!15}\textbf{2M} & \cellcolor{gray!15}\textbf{0.401}  \\ 
\midrule
\multirow{3}{*} {Weather} 
& PatchTST& 526M      & 0.598M          & 389M      & \textbf{2M}  & 0.225 \\
& SimMTM & 48865M       & 556M          & 259M      & 11M  & 0.227 \\ 
&\cellcolor{gray!15}\textbf{DeCoP} & \cellcolor{gray!15}\textbf{245M}      & \cellcolor{gray!15}0.463M          & \cellcolor{gray!15}\textbf{159M}      & \cellcolor{gray!15}2M & \cellcolor{gray!15}\textbf{0.223} \\\bottomrule
\end{tabular}
}
\caption{Comparison of FLOPs, parameters, and average MAE on the ETTh1 and weather dataset across different pre-training frameworks.}
\vspace{-3mm}
\label{table_flops}
\end{table}

\begin{table}[t]
    \centering
    \resizebox{0.48\textwidth}{!}{
    \begin{tabular}{c|cc|cc|cc|c}
        \toprule
        \multirow{2}{*}{$W_k$} & \multicolumn{2}{c|}{ETTh1} & \multicolumn{2}{c|}{ETTh2} & \multicolumn{2}{c|}{ETTm1}  & \multirow{2}{*}{Params} \\
        & MSE & MAE & MSE & MAE & MSE & MAE \\
        \midrule
        1,1   & 0.406 & 0.424 & 0.335 & 0.383 & 0.348 & 0.382  & 0.165M\\
        1,3   & 0.403 & 0.422 & 0.335 & 0.384 & 0.347 & 0.381  & 0.446M\\
        2,5   & \textbf{0.401} & \textbf{0.421} & \textbf{0.333} & \textbf{0.382} & 0.346 & 0.377 & 0.999M\\
        4,8   & 0.403 & 0.423 & 0.337 & 0.385 & \textbf{0.342} & \textbf{0.376} & 2.3M\\
        42,42 & 0.405 & 0.423 & 0.335 & 0.383 & 0.345 & 0.377  & 88.8M\\
        \bottomrule
    \end{tabular}
    }
    \caption{Controllable window sizes $W_k$ enable efficient pretraining by allowing flexible dependency modeling. A configuration around $(2, 5)$ achieves the best trade-off between accuracy and efficiency.}
    \label{table6}
    \vspace{-3mm}
\end{table}

\begin{figure}[t]
  \centering
   \includegraphics[width=1.0\linewidth]{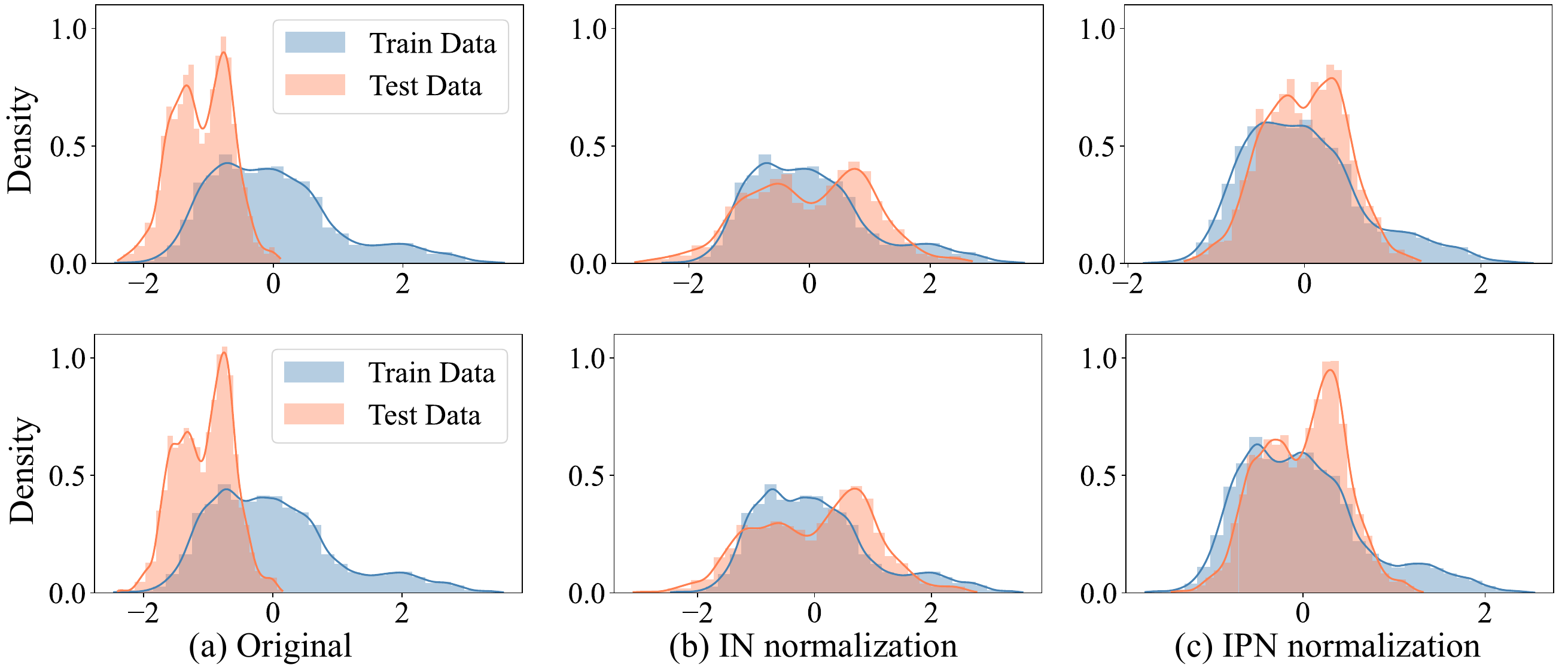}
   \caption{Comparison of data distributions before normalization (a), after IN (b), and after IPN(c) on ETTh1 and ETTm1. IPN preserves original semantic patterns such as peaks while better aligning train and test distributions.}
   \label{figure8}
   \vspace{-3mm}
\end{figure}

\textbf{The controllability of the DeCoP.}
A key challenge in time series pre-training is modeling temporal dependencies that evolve due to distribution shifts (\Cref{figure8}a) and multi-scale patterns, often resulting in noisy features and poor generalization. While existing approaches like PatchTST employ instance-level normalization (IN) to mitigate distribution shifts, we observe that IN tends to oversmooth patch-level variations, weakening semantic expressiveness (\Cref{figure8}b). In contrast, DeCoP explicitly addresses this challenge through controllable normalization. IPN adaptively normalizes both fine-grained patch-level statistics and coarse-grained instance-level distributions. This dual-level normalization allows the model to better preserve local temporal semantics while maintaining global statistical alignment. Compared to IN, IPN more effectively retains informative intra-patch variations (\Cref{figure8}c).

\begin{figure}[t]
  \centering
   \includegraphics[width=1.0\linewidth]{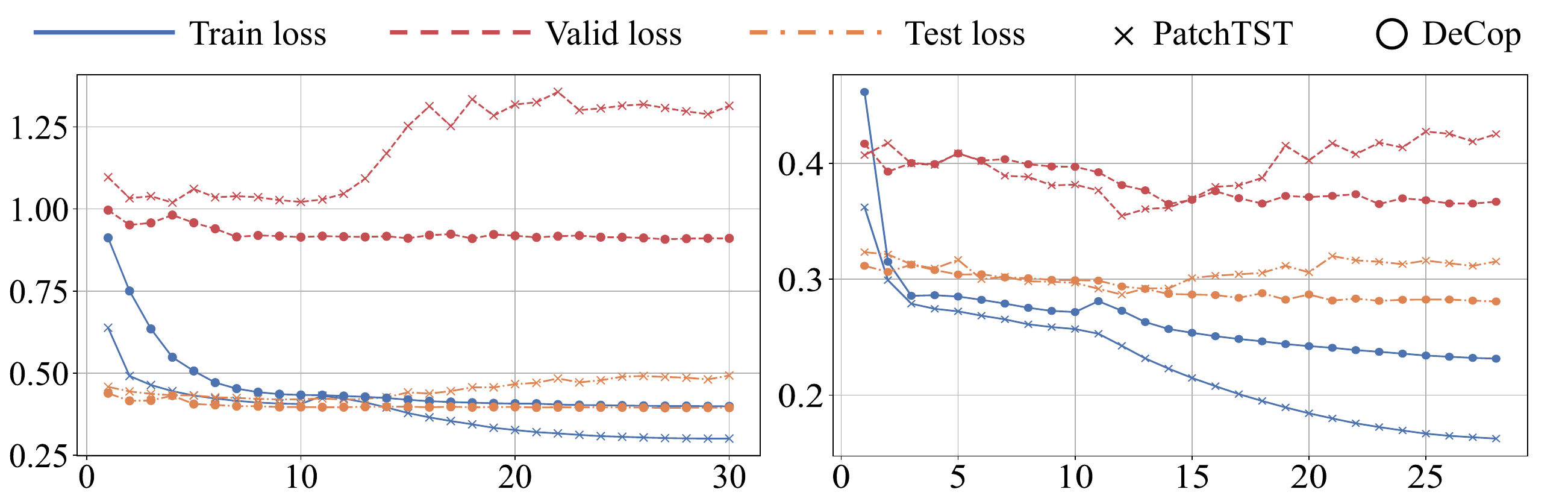}
   \caption{DeCoP achieves faster convergence and smaller train–val loss gaps on ETT datasets.}
   \label{train_loss}
   \vspace{-3mm}
\end{figure}

Additionally, our DCL method controllably encodes temporal structures through dynamic window grouping and encoding dependencies hierarchically. Unlike single scale attention in PatchTST, DCL explicitly constrains the temporal scope of dependency modeling, allowing the model to capture meaningful local patterns and gradually expand to global semantics. This controllable design reduces overfitting risks under distribution shifts by avoiding noisy or irrelevant dependencies. In \Cref{train_loss}, DeCoP converges faster and maintains a smaller gap between training and validation loss, demonstrating better generalization and reduced overfitting compared to PatchTST.

\textbf{Incorporating periodicity prior for efficient training.}
DeCoP outperforms prior methods such as PatchTST and SimMTM, achieving superior performance with fewer parameters through a controllable modeling mechanism. The DCL module captures dependencies between patches using variable window sizes, enabling alignment with periodic patterns in time series data. Empirically, we adopt a $(2, \ast)$ configuration to capture daily periodicity in hourly datasets with a patch size of 12, and a $(4, \ast)$ configuration to capture hourly periodicity in 10-minute datasets. In \Cref{table6}, The $(2, 5)$ setting achieves strong performance with only 999k parameters, highlighting the efficiency of DCL. 

\section{Conclusion}
This paper introduces \textbf{DeCoP}, a Dependency Controlled Pretraining framework that improve time series representation learning by explicitly modeling dynamic and multi-scale temporal dependencies. 
At the input level, IPN establishes a stable foundation by mitigating distribution shifts through instance-wise patch normalization while preserving fine-grained, patch-level information. At the latent representation level, DCL explicitly captures multi-scale dependencies through controllable receptive filed and ICM enhances global representation learning by incorporating time-invariant positive pairs.
DeCoP outperforms existing models with fewer parameters, highlighting the importance of dependency-controlled pre-training for dynamic time series. We hope that our work provides a strong baseline to future works on time series pretraining frameworks.

\bibliographystyle{ieee_fullname}

\begin{thebibliography}{10}\itemsep=-1pt

\bibitem{EPILEPSY}
Ralph~G Andrzejak, Klaus Lehnertz, Florian Mormann, Christoph Rieke, Peter David, and Christian~E Elger.
\newblock Indications of nonlinear deterministic and finite-dimensional structures in time series of brain electrical activity: Dependence on recording region and brain state.
\newblock {\em Physical Review E}, 2001.

\bibitem{deng2021graph}
Ailin Deng and Bryan Hooi.
\newblock Graph neural network-based anomaly detection in multivariate time series.
\newblock In {\em AAAI}, volume~35, pages 4027--4035, 2021.

\bibitem{devlin2018bert}
Jacob Devlin, Ming-Wei Chang, Kenton Lee, and Kristina Toutanova.
\newblock Bert: Pre-training of deep bidirectional transformers for language understanding.
\newblock In {\em Proceedings of the 2019 Conference of the North American Chapter of the Association for Computational Linguistics: Human Language Technologies (NAACL-HLT)}, 2019.

\bibitem{dong2024simmtm}
Jiaxiang Dong, Haixu Wu, Haoran Zhang, Li Zhang, Jianmin Wang, and Mingsheng Long.
\newblock Simmtm: A simple pre-training framework for masked time-series modeling.
\newblock In {\em NeurIPS}, volume~36, 2024.

\bibitem{ekambaram2020attention}
Vijay Ekambaram, Kushagra Manglik, Sumanta Mukherjee, Surya Shravan~Kumar Sajja, Satyam Dwivedi, and Vikas Raykar.
\newblock Attention based multi-modal new product sales time-series forecasting.
\newblock In {\em SIGKDD}, pages 3110--3118, 2020.

\bibitem{goswami2024moment}
M. Goswami, K. Szafer, A. Choudhry, Y. Cai, S. Li, and A. Dubrawski.
\newblock Moment: A family of open time-series foundation models.
\newblock {\em arXiv preprint}, 2024.

\bibitem{jin2023timellm}
M. Jin, S. Wang, L. Ma, Z. Chu, J.~Y. Zhang, X. Shi, and Q. Wen.
\newblock Time-llm: Time series forecasting by reprogramming large language models.
\newblock {\em arXiv preprint}, 2023.

\bibitem{kemp2000analysis-sleepeeg}
Bob Kemp, Aeilko~H Zwinderman, Bert Tuk, Hilbert~AC Kamphuisen, and Josefien~JL Oberye.
\newblock Analysis of a sleep-dependent neuronal feedback loop: the slow-wave microcontinuity of the eeg.
\newblock {\em TBME}, 2000.

\bibitem{kim2021reversible}
Taesung Kim, Jinhee Kim, Yunwon Tae, Cheonbok Park, Jang-Ho Choi, and Jaegul Choo.
\newblock Reversible instance normalization for accurate time-series forecasting against distribution shift.
\newblock In {\em ICLR}, 2021.

\bibitem{lee2023learning}
Seunghan Lee, Taeyoung Park, and Kibok Lee.
\newblock Learning to embed time series patches independently.
\newblock {\em arXiv preprint arXiv:2312.16427}, 2023.

\bibitem{FD-B}
Christian Lessmeier, James~Kuria Kimotho, Detmar Zimmer, and Walter Sextro.
\newblock Condition monitoring of bearing damage in electromechanical drive systems by using motor current signals of electric motors: A benchmark data set for data-driven classification.
\newblock In {\em PHM}, 2016.

\bibitem{li2023ti-TiMAE}
Zhe Li, Zhongwen Rao, Lujia Pan, Pengyun Wang, and Zenglin Xu.
\newblock Ti-mae: Self-supervised masked time series autoencoders.
\newblock {\em arXiv preprint arXiv:2301.08871}, 2023.

\bibitem{lin2024cyclenet}
Shengsheng Lin, Weiwei Lin, Xinyi Hu, Wentai Wu, Ruichao Mo, and Haocheng Zhong.
\newblock Cyclenet: enhancing time series forecasting through modeling periodic patterns.
\newblock {\em arXiv preprint arXiv:2409.18479}, 2024.

\bibitem{liu2023itransformer}
Yong Liu, Tengge Hu, Haoran Zhang, Haixu Wu, Shiyu Wang, Lintao Ma, and Mingsheng Long.
\newblock itransformer: Inverted transformers are effective for time series forecasting.
\newblock {\em arXiv preprint arXiv:2310.06625}, 2023.

\bibitem{liu2022non}
Yong Liu, Haixu Wu, Jianmin Wang, and Mingsheng Long.
\newblock Non-stationary transformers: Exploring the stationarity in time series forecasting.
\newblock In {\em NeurIPS}, volume~35, pages 9881--9893, 2022.

\bibitem{liu2024timer}
Y. Liu, H. Zhang, C. Li, X. Huang, J. Wang, and M. Long.
\newblock Timer: Generative pre-trained transformers are large time series models.
\newblock In {\em ICML}, 2024.

\bibitem{nie2022time}
Yuqi Nie, Nam~H Nguyen, Phanwadee Sinthong, and Jayant Kalagnanam.
\newblock A time series is worth 64 words: Long-term forecasting with transformers.
\newblock {\em arXiv preprint arXiv:2211.14730}, 2022.

\bibitem{EMG}
PhysioToolkit PhysioBank.
\newblock Physionet: components of a new research resource for complex physiologic signals.
\newblock {\em Circulation}, 2000.

\bibitem{rasul2023lag}
Kashif Rasul, Arjun Ashok, Andrew~Robert Williams, Arian Khorasani, George Adamopoulos, Rishika Bhagwatkar, Marin Bilo{\v{s}}, Hena Ghonia, Nadhir Hassen, Anderson Schneider, et~al.
\newblock Lag-llama: Towards foundation models for time series forecasting.
\newblock In {\em R0-FoMo: Robustness of Few-shot and Zero-shot Learning in Large Foundation Models}, 2023.

\bibitem{ecldata}
{UCI}.
\newblock {UCI Electricity Load Time Series Dataset}.
\newblock \url{https://archive.ics.uci.edu/ml/datasets/ElectricityLoadDiagrams20112014}, 2021.

\bibitem{vaswani2017attention}
Ashish Vaswani, Noam Shazeer, Niki Parmar, Jakob Uszkoreit, Llion Jones, Aidan~N. Gomez, {\L}ukasz Kaiser, and Illia Polosukhin.
\newblock Attention is all you need.
\newblock In {\em NeurIPS}, volume~30, 2017.

\bibitem{wang2024timemixer}
Shiyu Wang, Haixu Wu, Xiaoming Shi, Tengge Hu, Huakun Luo, Lintao Ma, James~Y Zhang, and Jun Zhou.
\newblock Timemixer: Decomposable multiscale mixing for time series forecasting.
\newblock {\em arXiv preprint arXiv:2405.14616}, 2024.

\bibitem{wang2022learning}
Zhiyuan Wang, Xovee Xu, Weifeng Zhang, Goce Trajcevski, Ting Zhong, and Fan Zhou.
\newblock Learning latent seasonal-trend representations for time series forecasting.
\newblock In {\em NeurIPS}, 2022.

\bibitem{weatherdata}
{Wetterstation}.
\newblock {Weather Dataset}.
\newblock \url{https://www.bgc-jena.mpg.de/wetter/}, 2021.

\bibitem{woo2024unified}
G. Woo, C. Liu, A. Kumar, C. Xiong, S. Savarese, and D. Sahoo.
\newblock Unified training of universal time series forecasting transformers.
\newblock {\em arXiv preprint}, 2024.

\bibitem{woo2022cost}
Gerald Woo, Chenghao Liu, Doyen Sahoo, Akshat Kumar, and Steven Hoi.
\newblock Cost: Contrastive learning of disentangled seasonal-trend representations for time series forecasting.
\newblock In {\em ICLR}, 2022.

\bibitem{wu2022timesnet}
Haixu Wu, Tengge Hu, Yong Liu, Hang Zhou, Jianmin Wang, and Mingsheng Long.
\newblock Timesnet: Temporal 2d-variation modeling for general time series analysis.
\newblock In {\em ICLR}. arXivpreprint, 2023.

\bibitem{wu2021autoformer}
Haixu Wu, Jiehui Xu, Jianmin Wang, and Mingsheng Long.
\newblock Autoformer: Decomposition transformers with auto-correlation for long-term series forecasting.
\newblock In {\em NeurIPS}, volume~34, pages 22419--22430, 2021.

\bibitem{wu2023}
H. Wu, H. Zhou, M. Long, and J. Wang.
\newblock Interpretable weather forecasting for worldwide stations with a unified deep model.
\newblock {\em Nature Machine Intelligence}, 5(6):602--611, 2023.

\bibitem{yue2022ts2vec}
Zhanwei Yue, Yiqun Wang, Jinghua Duan, Tao Yang, Chen Huang, Yunhai Tong, and Bo Xu.
\newblock Ts2vec: Towards universal representation of time series.
\newblock In {\em AAAI}, 2022.

\bibitem{zeng2023transformers}
Ailing Zeng, Muxi Chen, Lei Zhang, and Qiang Xu.
\newblock Are transformers effective for time series forecasting?
\newblock In {\em AAAI}, 2023.

\bibitem{zerveas2021transformer}
George Zerveas, Srideepika Jayaraman, Dhaval Patel, Anuradha Bhamidipaty, and Carsten Eickhoff.
\newblock A transformer-based framework for multivariate time series representation learning.
\newblock In {\em SIGKDD}, 2021.

\bibitem{zhang2022self}
Xiang Zhang, Ziyuan Zhao, Theodoros Tsiligkaridis, and Marinka Zitnik.
\newblock Self-supervised contrastive pre-training for time series via time-frequency consistency.
\newblock {\em NeurIPS}, 35:3988--4003, 2022.

\bibitem{Zhang2022-TF-C}
Xiang Zhang, Ziyuan Zhao, Theodoros Tsiligkaridis, and Marinka Zitnik.
\newblock Self-supervised contrastive pre-training for time series via time-frequency consistency.
\newblock In {\em NeurIPS}, 2022.

\bibitem{zhou2021informer}
Haoyi Zhou, Shanghang Zhang, Jieqi Peng, Shuai Zhang, Jianxin Li, Hui Xiong, and Wancai Zhang.
\newblock Informer: Beyond efficient transformer for long sequence time-series forecasting.
\newblock In {\em AAAI}, pages 11106--11115, 2021.

\bibitem{zhou2022fedformer}
Tian Zhou, Ziqing Ma, Qingsong Wen, Xue Wang, Liang Sun, and Rong Jin.
\newblock Fedformer: Frequency enhanced decomposed transformer for long-term series forecasting.
\newblock In {\em ICML}, pages 27268--27286. PMLR, 2022.

\end{thebibliography}
\end{document}